\documentclass[letterpaper]{article} 
\usepackage[preprint]{aaai2027}  

\usepackage[hyphens]{url}  
\usepackage{graphicx} 
\usepackage{natbib}  
\usepackage{caption} 
\usepackage{algorithm}

\usepackage{amsfonts}
\usepackage{amsmath}
\usepackage{booktabs}
\usepackage{tabularx}
\usepackage{multirow}
\usepackage{makecell}
\usepackage{graphicx}
\usepackage{algpseudocode}
\usepackage{amssymb}

\usepackage{listings}
\DeclareCaptionStyle{ruled}{labelfont=normalfont,labelsep=colon,strut=off} 
\floatstyle{ruled}
\newfloat{listing}{tb}{lst}{}
\floatname{listing}{Listing}

\usepackage{booktabs}

\title{Learning Latent Reasoning Traces for Scalar Reward Models End-to-End}
\author{
    Sanwoo Lee\textsuperscript{\rm 1,2,3}\internship,
    Clive Bai\textsuperscript{\rm 3},
    Hsiu-Yuan Huang\textsuperscript{\rm 1,2,3}\internship,
    Kun Liang\textsuperscript{\rm 1,2,3}\internship,
    Weijie Liu\textsuperscript{\rm 3},
    Yunfang Wu\textsuperscript{\rm 1,2}\corresponding
}
\affiliations{
    \textsuperscript{\rm 1}National Key Laboratory for Multimedia Information Processing, Peking University\\
    \textsuperscript{\rm 2}School of Computer Science, Peking University  \quad \textsuperscript{\rm 3}LLM Department, Tencent \\
    sanwoo@pku.edu.cn \quad wuyf@pku.edu.cn 
}

\begin{document}

\maketitle

\begin{abstract}

Reward models (RMs) are central to aligning large language models with human preferences via reinforcement learning. Although traditional scalar RMs enable efficient and probabilistic reward modeling, they rely on superficial cues that fail to generalize to complex or out-of-distribution (OOD) tasks. Conversely, generative RMs leverage extensive reasoning to improve robustness on challenging tasks, but their natural language-based scores lack the numerical flexibility and probabilistic interpretability that scalar RMs offer. While recent approaches combine both paradigms through off-policy multi-task learning, such parallel optimization does not guarantee that generated reasoning traces actively align with or benefit downstream scalar reward prediction. To address this mismatch, we propose \textbf{LatentRM}, a reward modeling framework that learns intermediate reasoning traces as discrete latent variables to explicitly maximize the likelihood of downstream scalar rewards. Through on-policy optimization of the latent reasoning space end-to-end, LatentRM tightly couples deep reasoning-based evaluation with precise scoring. Extensive validations on in-distribution and OOD datasets and RLHF show that LatentRM outperforms scalar, generative, and hybrid RMs on preference modeling and policy alignment across tasks ranging from open-ended conversation to complex reasoning.

\end{abstract}

\section{Introduction}

Reward models (RMs) have shown great promise in aligning large language models (LLMs) with human preference \citep{ouyang22training, bai2022training}. Central to the reinforcement learning from human feedback (RLHF) process, RMs offer an efficient and scalable proxy for human preference, improving LLMs on a wide range of tasks \citep{nips223fine} beyond domains where rule-based verifiable rewards are readily available. Therefore, a key challenge in developing RMs lies in establishing an accurate and faithful proxy robust to potential distribution shifts in prompt and response. 

\begin{figure}[t]
    \centering
    \includegraphics[width=\linewidth]{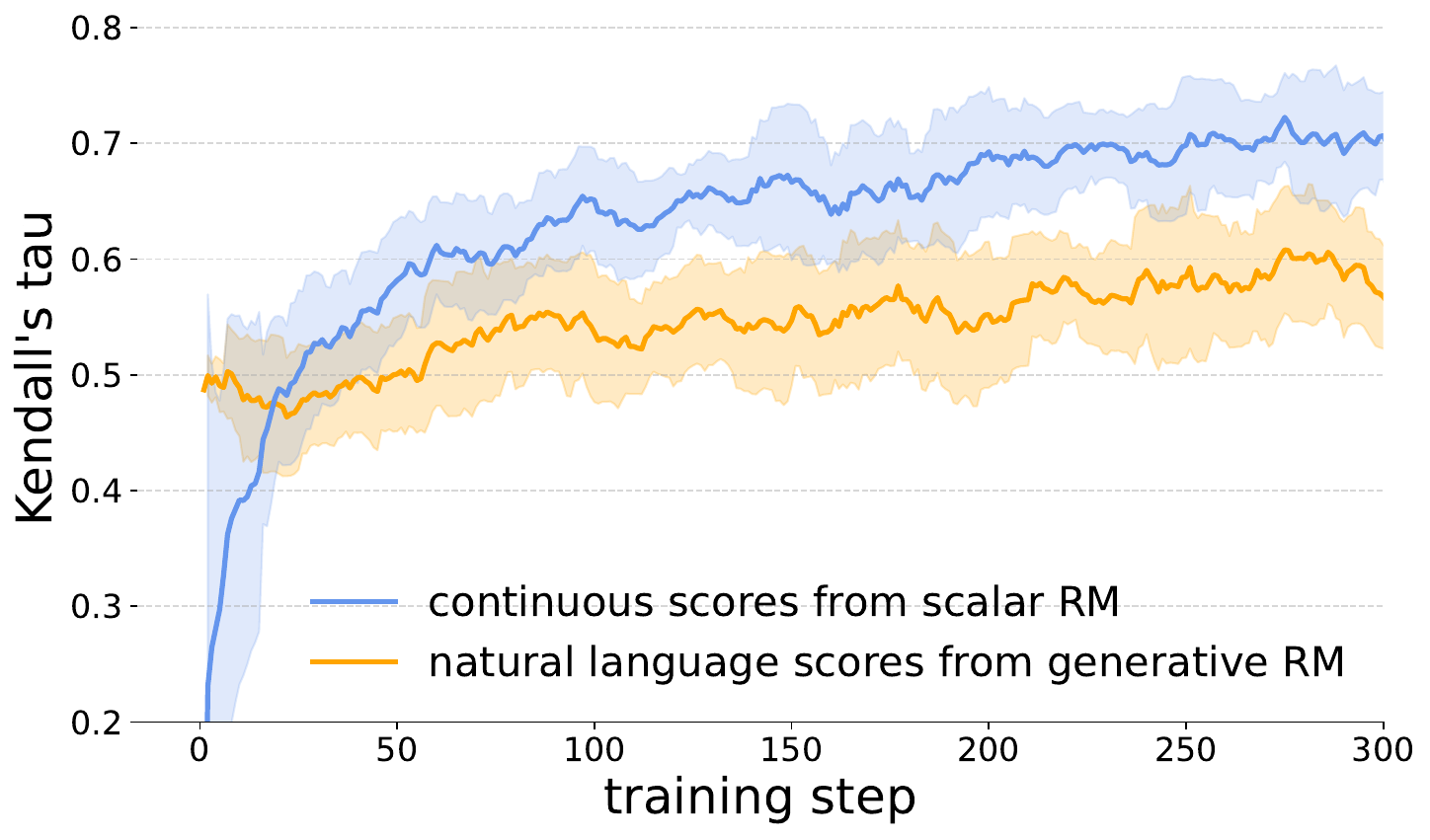}
    \caption{Rank correlation coefficient against groundtruth ranking of scalar RM and generative RM, measured over 1 epoch training on the training set. Both models are initialized from Qwen3-4B-Instruct-2507 \citep{qwen3technicalreport}, where acalar RM is trained to predict under the same prefix context as generative RM. }
    \label{fig:pilot_study}
\end{figure}

RMs are typically designed as scalar RM which places a linear head on top of an LLM backbone and learns from preference rankings, to assign a scalar score for a pair of prompt and response \citep{ziegler2019fine, ouyang22training}. However, scalar RMs struggle with distribution shift \citep{icml23scaling} and reasoning-intensive prompts \citep{nips25think} as they often overfit to superficial patterns in the training data. Later works have mitigated these issues with generative RMs that follows LLM-as-Judge paradigm and invokes deep reasoning traces to substantiate the final rankings or scores \citep{chen2026rmr, whitehouse2026j}, yet the scores in natural language lack the flexibility and probabilistic interpretation that scalar RMs naturally offer. As Figure~\ref{fig:pilot_study} illustrates, when optimizing a generative RM for Kendall's $\tau$ against groundtruth ranking, a scalar RM which shares exactly same prefix context as generative RM for scoring achieves markedly superior performance. This shows that the flexibility of scalar RM translates to more accurate preference modeling when backed up by the reasoning as additional context, suggesting the need for an effective interplay between these two paradigms.

Despite recent efforts to enhance scalar RMs with reasoning, seamlessly combining the two remains challenging. An initial work explored appending critiques generated by  external teacher LLMs to the input of a scalar RM \citep{ye-etal-2025-improving}, demonstrating consistent gains across teacher models. Subsequent works resolved the reliance on teachers at inference time by jointly training the scalar RM on critique distillation and preference learning \citep{ankner2024critique}, or removed external teachers altogether by leveraging self-filtered critiques for distillation instead \citep{yu-etal-2025-self}. However, fine-tuning from static critiques, even the self-generated and filtered ones, incurs a mismatch between training and inference, since the reasoning the model generates becomes increasingly off-policy as training progresses. Most importantly, the enhancement of reasoning quality relies on ad-hoc heuristics, such as calling strong LLMs and self-refinement, leaving their contribution to the downstream reward prediction largely unclear. This underscores the need for a task-aligned objective that refines the reasoning trace explicitly for assisting the scalar RM.

In this work, we propose \textbf{LatentRM}, a scalar reward model with a generator that learns to shape its reasoning for a \textit{unified} goal of maximizing the likelihood of its downstream scalar reward. Specifically, we treat the reasoning as a discrete latent variable $z$ bridging the input $x$ and preference label $y$ in a conditional generative model. In this framework, we optimize for the data likelihood $\log p(y|z)$ under Plackett-Luce model \citep{luce1959individual}, by maximizing its lower bound with a simple and affordable choice of variational posterior. We transform this objective into an end-to-end joint training procedure where both scalar RM and generator are updated on the on-policy critiques from the generator. Through this design, LatentRM achieves a tight interplay between generator and scalar RM for the ultimate goal of accurate preference learning.  Experiments on in-distribution (ID) and out-of-distribution (OOD) datasets, as well as RLHF alignment, demonstrate that LatentRM outperforms scalar RM, generative RM and a multi-task hybrid RM where the generator has strong yet separate learning objective from the scalar RM. 

In essence, we articulate our contributions as follows:
\begin{itemize}
\item We present LatentRM, an end-to-end framework that treats reasoning as a latent variable to directly optimize downstream scalar RM's preference modeling, which replaces manually designed reward.
\item We devise an on-policy training procedure derived from the data likelihood ELBO, co-optimizing the generator and scalar RM that removes training-inference mismatch.
\item LatentRM showcases superior ID and OOD performance over scalar, generative and hybrid RMs on diverse benchmarks without introducing hyperparameter overhead.
\end{itemize}

\section{Related Work}

\paragraph{Reward Models} 

Reward models have been central in aligning LLMs during on-policy training \citep{ouyang22training} as well as test-time generation \cite{frick2025how}. Traditional RMs are typically scalar RMs trained with Bradley-Terry loss \citep{Bradley1952} over pairwise preference data \cite{christ2017deep, bai2022training}. Scalar RMs had often suffered from distribution shifts \citep{miao24inform} and consume massive datasets to generalize \citep{ye-etal-2025-improving}, which prompted active research on fine-tuning generative RMs to elicit reasoning prior to rewards in textual format \citep{kim2024prometheus, li2024autoj}. This approach has been extended to leveraging reinforcement learning to scale reasoning towards deeper depth \citep{chen2026rmr, whitehouse2026j}. Concurrently, hybrid architectures generate a rationale and pass it to a scalar RM for reward prediction, fine-tuning both on preference labels alongside rationales distillted from stronger LLMs \citep{ye-etal-2025-improving, ankner2024critique} or self-refinement process \citep{yu-etal-2025-self}. Nevertheless, few is known about the link between refining reasoning and optimizing downstream scalar RM, which we aim to address by a unified objective.

\paragraph{Latent Variable View of Reasoning}
Chain-of-thought reasoning serves as a key driver for improving LLM performance on downstream applications. Early foundational work by \citet{lei-etal-2016-rationalizing} trained an extractive rationale generator by treating rationales as latent variables that maximize a downstream discriminator's predictive accuracy. Building on this perspective, STaR \citep{zelikman2022star} iteratively fine-tuned models on self-generated rationales filtered by binary verifiers, in effect approximating a policy gradient objective over latent reasoning traces. This approach was later extended using Markov-chain Monte Carlo sampling to generate rationales that align with both the generator and the ground-truth answer \citep{phan23training}. Most recently, \citet{tang25beyond} introduced a fully on-policy RL method to learn latent reasoning for semi-verifiable tasks, such as mathematical proofs. However, leveraging latent variable reasoning frameworks for reward modeling remains largely unexplored. To our knowledge, LatentRM is the first work to formalize reasoning as a latent variable in a generator–scalar RM architecture.

\section{Method}

\subsection{Problem Statement}

We consider a general reward modeling problem where the input $x$ consists of a prompt $x_p$ and $k$ candidate responses $\{x_{r_i}\}_{i=1}^{k}$. For the sake of generality, we let $k \geq 1$ and posit that   $k$ may vary across prompts, which subsumes canonical pointwise and pairwise reward modeling setup as special cases. The reward model then predicts $k$ real-valued rewards $s= (s_1, s_2, ..., s_k) \in \mathbb{R}^{k}$ for the responses. The goal is to train a reward model whose predicted ranking induced by $s$ is aligned with the human preference ranking $y$ over the responses.

\subsection{Model Architecture}

In this paper, we introduce a generator-discriminator architecture for the reward model. In particular, the generator LLM parametrized by $\theta \in \mathbb{R}^{d_{\theta}}$ samples a chain-of-thought reasoning $z \sim p_{\theta}(z|x)$ given $x$. Then the discriminator scalar RM parametrized by $\varphi \in \mathbb{R}^{d_\varphi}$ outputs rewards $s \in \mathbb{R}^{k}$ given $x$ and $z$.  This design allows the discriminator to leverage ample evidence supporting the final decision in the reasoning, while ensuring flexible scoring that generative RMs lack. While the scalar RM may be of different architecture from the generator, we simply initialize the scalar RM as a copy of the generator with the language modeling head replaced by a scalar head randomly parametrized by $w \in \mathbb{R}^{h}$. To induce a distribution over the the listwise rankings $p_{\varphi}$, we use Plackett-Luce model \citep{luce1959individual}. For a strict total ordering $y=(y_1 \succ y_2 \succ ... \succ y_k)$, its likelihood is defined as:
\begin{equation}
\label{eq:plackett-luce}
    p_{\varphi}(y|x,z) = \prod_{i=1}^{k} \frac{\exp(s_{y_i})}{\sum_{j=i}^{k} \exp(s_{y_j})}
\end{equation}
which reduces to the standard Bradley-Terry model when $k=2$. When ties are present in $y$, we sum over the likelihoods of all strict total orderings $\pi = (\pi_1 \succ \pi_2 \succ ... \succ \pi_k) \in \Omega(y)$ consistent with the weak ordering $y$ containing ties.

\begin{figure}[ht]
    \centering
    \includegraphics[width=\linewidth]{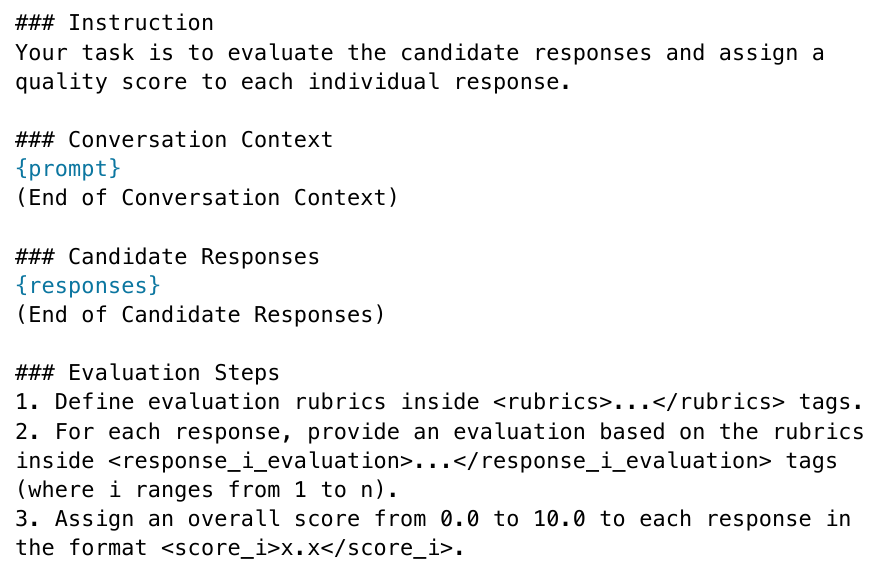}
    \caption{A condensed prompt template for our listwise generator. Full template is available at Figure~\ref{fig:full_template} in Appendix.}
    \label{fig:template_summary}
\end{figure}

To implement the generator-discriminator inference flow, we organize our prompt template as in Figure~\ref{fig:template_summary} that wraps the prompt and candidate responses as input $x$ for reasoning generation. The reasoning trace complies with the format requirements of the template, yielding a rubric-based verbal evaluation followed by an overall score for each response in format  \texttt{<score\_i>x.x</score\_i>} ($i=1, 2,...,k$). We then feed the concatenated input and reasoning trace $(x, z)$ into the scalar RM, and extract the last-layer hidden states at the $k$ token positions immediately preceding the generated score values, namely the last tokens in \texttt{<score\_i>}. Stacking these hidden states column-wise yields
\begin{equation}
    H = [h_1, \ldots, h_k] \in \mathbb{R}^{h \times k}.
\end{equation}
We then apply a shared scalar head $w \in \mathbb{R}^{h}$ to obtain the rewards as $s = w^{\top} H$.

\begin{figure}[t]
    \centering
    \includegraphics[width=\linewidth]{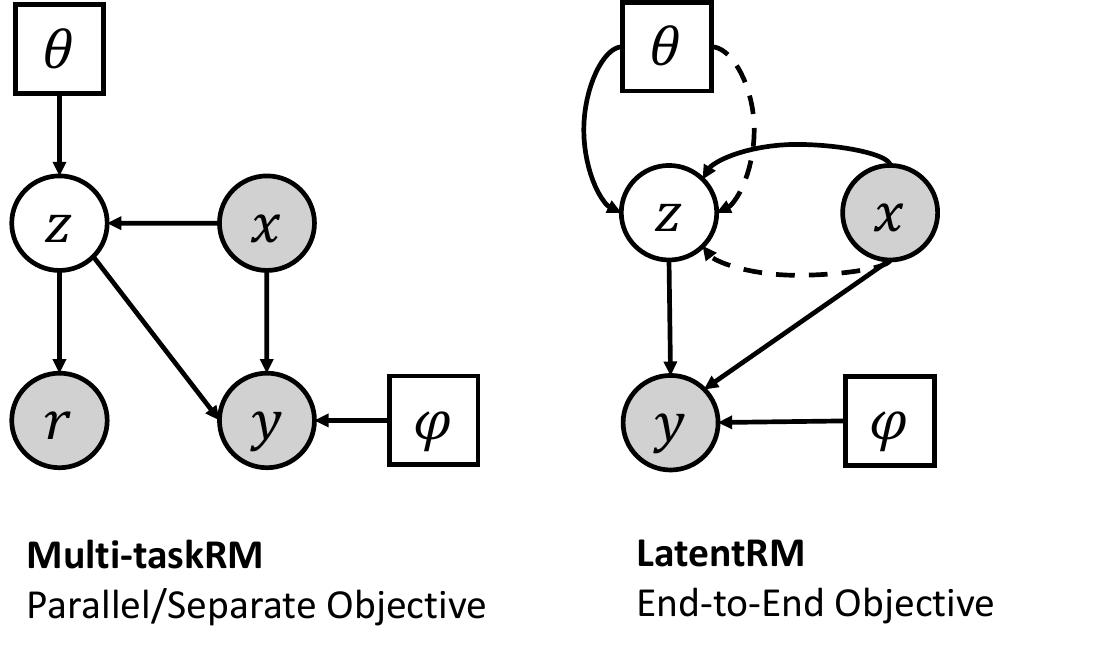}
    \caption{Graphical models illustrating differences between MultitaskRM (baseline) and LatentRM. Solid line denotes generative model and dashed line denotes inference models. Circles are random variables and squares are parameters, where shaded circles represent observed variables.}
    \label{fig:graphical_model}
\end{figure}

\subsection{Learning to Reason as Latent Variables}

A straightforward approach for generator-discriminator architectures is co-training both components in parallel using each component's own objective. Generators are widely trained with binary correctness reward \citep{nips25think, whitehouse2026j} while discriminators are predominantly trained with Bradley-Terry loss \citep{bai2022training}, which generalizes to Kendall's $\tau$ rank correlation reward and Plackett-Luce loss under listwise setup. However, these objectives were originally formulated for individual components in isolation. Combining them via naive multi-task learning, which we denote as \textbf{Multitask RM}, does not guarantee optimal accuracy of downstream reward, as maximizing surrogate rewards like Kendall's $\tau$ serves only as an imperfect proxy for maximizing the true log-likelihood $\log p_{\varphi}(y|x,z)$ (Eq.~\ref{eq:plackett-luce}).

To address this issue, we establish an end-to-end objective by casting CoT reasoning $z$ as a discrete latent variable that conditions the scalar RM (discriminator) for better prediction. This is equivalent to formalizing the architecture as a conditional generative model \citep{nips15cvae}, where given $x$, the latent variable $z$ is drawn from the prior $p_\theta(z|x)$ and the output $y$ is drawn from the decoder network $p_\varphi(y|x,z)$. Under the generative model, our goal is to jointly train the parameters $(\theta, \varphi)$ to maximize the conditional log-likelihood $\log p(y|x)$, marginalizing over the latent reasoning:
\begin{equation}
\label{eq:obj}
    \max_{\theta, \varphi} \log p(y|x) = \max_{\theta, \varphi} \log \mathbb{E}_{z\sim p_\theta(z|x)} \left[ p_{\varphi}(y|x,z) \right]
\end{equation}
Directly optimizing Eq.~\ref{eq:obj} is intractable because its gradient requires expectations over the true posterior $p(z|x,y)$. The standard variational approach optimizes its evidence lower bound (ELBO) by introducing an inference network $q_\phi(z|x, y)$ as an approximation to the posterior: 
\begin{equation}
\begin{split}
    \log p(y|x) = \, &\mathbb{E}_{q_\phi(z|x,y)} \left[ \log \frac{p_\theta(z|x)p_\varphi(y|x,z)}{q_\phi(z|x,y)} \right] \\
    &+ D_{KL} \left(q_\phi(z|x,y) \parallel p(z|x,y)\right) \\
    \ge \, &\mathbb{E}_{q_\phi(z|x,y)} \left[ \log p_\varphi(y|x,z) \right] \\
    &- D_{KL} \left(q_\phi(z|x,y) \parallel p_\theta(z|x)\right) = \mathcal{L}_{\text{ELBO}}
\end{split}
\end{equation}
However, optimization of parameterized $q_{\phi}(z|x,y)$ requires training a second LLM conditioned on $y$. This adds substantial computational cost and is conceptually counter-intuitive, as $q_{\phi}(z|x,y)$ risks generating spurious justification solely to match any given label rather than discovering authentic task logic. Therefore, we simply choose the inference network to be shared by the prior, i.e., $q_{\phi}(z|x,y)=p_{\theta}(z|x)$.  By Jensen's inequality, this choice still recovers a valid lower bound on the true marginal likelihood. 
The resulting optimization objective thus simplifies to:
\begin{equation}
\label{eq:lb}
\max_{\theta, \varphi} \mathbb{E}_{z\sim p_\theta(z|x)} \left[ \log p_{\varphi}(y|x,z) \right]
\end{equation}
Figure~\ref{fig:graphical_model} compares graphical models for MultitaskRM and LatentRM. LatentRM leverage solely $x$ and $y$ to jointly supervise $\theta$ and $\varphi$, whereas MultitaskRM introduces an auxiliary reward $r$ to train $\theta$ in parallel with $\varphi$.

\subsection{Gradient of the Lower Bound and its Estimator}
To switch from the objective in Eq.~\ref{eq:lb} to practical training updates,  we first derive the gradients of $\mathcal{L}_{\text{ELBO}}$ w.r.t. $\theta$ and $\varphi$ as follows:
\begin{align}
& \nabla_\varphi \mathcal{L}_{\text{ELBO}} \nonumber \\
& = \mathbb{E}_{z \sim p_\theta} \left[ \nabla_\varphi \log p_\varphi(y|x, z) \right] \label{eq:grad_phi_pl} \\
& = \mathbb{E}_{z \sim p_\theta} \left[ \sum_{i=1}^k \left( 1 - \sum_{j=1}^{i} \frac{\exp(s_{y_i})}{\sum_{l=j}^{k} \exp(s_{y_l})} \right) \nabla_{\varphi} s_{y_i}(x, z) \right] \nonumber \\
& \nabla_\theta \mathcal{L}_{\text{ELBO}} \nonumber \\
& = \mathbb{E}_{z \sim p_\theta} \left[ \log p_\varphi(y|x, z) \nabla_\theta \log p_\theta(z|x) \right] \label{eq:grad_theta_pl} 
\end{align}
Eq.~\ref{eq:grad_phi_pl} and \ref{eq:grad_theta_pl} reveal LatentRM's optimization mechanism. For a given $(x,y)$, the scalar RM is updated to increase the expected log-likelihood of $y$ under the current reasoning distribution. Under the Plackett-Luce model, this amounts to correcting each score $s_{y_i}$ in proportion to the scalar RM's remaining uncertainty about $y_i$'s rank, a gap that vanishes once the it ranks $y_i$ correctly without ambiguitiy in scores. Meanwhile, the generator is updated to favor reasoning traces $z$ that increase the discriminator's log-likelihood of observing $y$.

We present the training procedure of LatentRM in Algorithm~\ref{alg:training} with the  empirical estimators of the gradients. The generator samples $m$ rollouts (i.e., reasoning) per instance in a batch, then the scalar RM is updated with standard supervised learning. Since sampling the rollout $z$ is not differentiable, we update the generator using REINFORCE \citep{sutton99reinforce}, treating the log-likelihood $\log p_\varphi(y^{(i)} | x^{(i)}, z^{(i,j)})$ as the reward signal. To reduce the variance of this policy gradient estimator, we compute a baseline $b^{(i)}$ as the average reward across the $m$ rollouts for a given prompt, subtracting it from each rollout's reward to obtain the advantage $A^{(i,j)}$. For rollouts violating format requirements, we compute advantages solely over valid rollouts and assign invalid ones an advantage of $\min_{j \in \mathcal{V}(i)} A^{(i,j)} -1$ where $\mathcal{V}(i)$ is the set of valid rollouts. If $\mathcal{V}(i)=\varnothing$, every rollout receives an advantage of $-1$. Overall, $\theta$ and $\varphi$ are updated jointly at every iteration with their respective gradients. We observe such joint learning is effective and stable, not requiring warmup steps for scalar RM.

\begin{algorithm}[t]
\caption{LatentRM Training}
\label{alg:training}
\begin{algorithmic}[1]
    
\Require Training set $\mathcal{D}$, initial parameters $\theta_0, \varphi_0$, number of rollouts per prompt $m$, batch size $B$, learning rates $\eta_\theta, \eta_\varphi$
\State Initialize $\theta \leftarrow \theta_0$, $\varphi \leftarrow \varphi_0$
\For{each training step}
    \State Sample a mini-batch $\{(x^{(i)}, y^{(i)})\}_{i=1}^{B} \sim \mathcal{D}$
    \For{$i=1, \dots, B$}
        \State Sample $m$ rollouts $\{z^{(i,j)}\}_{j=1}^{m} \sim p_\theta(z | x^{(i)})$
        \State Compute $R^{(i,j)} \leftarrow \log p_\varphi(y^{(i)} | x^{(i)}, z^{(i,j)})$
        \State Compute baseline $b^{(i)} \leftarrow \frac{1}{m} \sum_{j=1}^{m} R^{(i,j)} $ 
        \State Compute advantage $A^{(i,j)} \leftarrow R^{(i,j)} - b^{(i)}$
    \EndFor
    \State $\varphi \leftarrow \varphi + \eta_{\varphi} \cdot \frac{1}{Bm} \sum_{i,j} \nabla_{\varphi} \log  p_\varphi(y^{(i)} | x^{(i)}, z^{(i, j)})$ 
    \State $\theta \leftarrow \theta + \eta_{\theta} \cdot \frac{1}{Bm} \sum_{i,j} A^{(i,j)} \nabla_{\theta} \log  p_\theta(z^{(i,j)} | x^{(i)})$ 
\EndFor
\State \Return $\theta$, $\varphi$
\end{algorithmic}
\end{algorithm}

\section{Experiments}

\subsection{Experimental Setup}
 We validate LatentRM's effectiveness on ID test set and several OOD benchmarks for evaluating reward models. In addition, we conduct RLHF experiments with LatentRM and baseline RMs to explore their utility on inducing an aligned LLM. Throughout the experiments, we use Qwen3-4B-Instruct\footnote{\url{https://huggingface.co/Qwen/Qwen3-4B-Instruct-2507}}~\citep{qwen3technicalreport} as the backbone model across LatentRM and baselines.

\begin{table}[thbp]
\centering
\caption{Dataset statistics. \textbf{\# Resp.}: number of responses per prompt; \textbf{\# Init.}/\textbf{\# Test.}/\textbf{\# Train}: sizes of the initial pool, test, and filtered training set. Dataset names are abbreviated by their initials.}
\label{tab:dataset_stats}
\setlength{\tabcolsep}{4pt}
\small
\begin{tabular}{llcccc}
\toprule
\textbf{Dataset} & \textbf{Domain} & \textbf{\# Resp.} & \textbf{\# Init.} & \textbf{\# Test.} & \textbf{\# Train} \\
\midrule
UF & General & 4 & 28,000 & 1680 & 15,120 \\
OMR & Mathematics & 2--8 & 28,000 & 1679 & 15,121 \\
HS3 & STEM\&Code & 2 & 8,000 & 480 & 4,320 \\
WG & Safety & 2 & 8,000 & 480 & 4,320 \\
OB & Adversarial & 2 & 8,000 & 480 & 4,320 \\
\midrule
Total & -- & -- & 80,000 & 4799 & 43,201 \\
\bottomrule
\end{tabular}
\end{table}

\paragraph{Datasets} To cover diverse domains, we construct our training data pool by randomly sampling preference-annotated data from UltraFeedback~\citep{24ultrafb} (28K), OpenMathReasoning~\citep{moshkov2025aimo} (28K), Helpsteer3~\citep{wang2026helpsteerpreference} (STEM and Code subsets, 8K), WildGuard~\citep{han2026wildguard} (adversarial subset, 8K), and OffsetBias~\citep{park-etal-2024-offsetbias} (8K). Dataset statistics are listed in Table~\ref{tab:dataset_stats}.

\begin{table*}[ht]
\centering
\small
\renewcommand{\arraystretch}{1.3}
\setlength{\tabcolsep}{0pt}
\begin{tabular*}{\textwidth}{@{\extracolsep{\fill}} ll ccccc cc @{}}
\toprule
 & & \textbf{General} & \textbf{Math} & \textbf{STEM \& Code} & \textbf{Safety} & \textbf{Adversarial} & \multicolumn{2}{c}{\textbf{Averages}} \\
\cmidrule(lr){3-3} \cmidrule(lr){4-4} \cmidrule(lr){5-5} \cmidrule(lr){6-6} \cmidrule(lr){7-7} \cmidrule(lr){8-9}
\textbf{Metric} & \textbf{Method} & \textbf{UltraFeedback} & \textbf{OMR} & \textbf{Helpsteer3} & \textbf{Wildguard} & \textbf{OffsetBias} & \textbf{Micro} & \textbf{Macro} \\
\midrule
\multirow{3}{*}{Log-likelihood } 
 & ScalarRM & -1.986 & -1.154 & -0.436 & \textbf{-0.109} & -0.221 & -1.175 & -0.781 \\
 & MultitaskRM & -1.917 & -0.901 & -0.379 & -0.142 & -0.153 & -1.053 & -0.698 \\
 & LatentRM & \textbf{-1.899} & \textbf{-0.840} & \textbf{-0.373} & -0.199 & \textbf{-0.149} & \textbf{-1.031} & \textbf{-0.692} \\
\midrule
\multirow{5}{*}{Kendall's $\tau$} 
 & ScalarRM & 0.649 & 0.606 & 0.600 & \textbf{0.929} & 0.807 & 0.673 & 0.718 \\
 & GenRM w/o RL & 0.592 & 0.398 & 0.506 & 0.496 & 0.676 & 0.497 & 0.534 \\
 & GenerativeRM & 0.586 & 0.541 & 0.470 & 0.758 & 0.702 & 0.588 & 0.611 \\
 & MultitaskRM & 0.655 & 0.677 & 0.706 & 0.875 & 0.866 & 0.706 & 0.756 \\
 & LatentRM & \textbf{0.662} & \textbf{0.685} & \textbf{0.712} & 0.850 & \textbf{0.887} & \textbf{0.712} & \textbf{0.759} \\
\bottomrule
\end{tabular*}
\caption{ Evaluation results on ID test sets, measured by log-likelihood $p_{\varphi}(y|x,z)$ and Kendall's tau. Log-likelihood is not applicable for GenerativeRMs as scores are in textual format. The best result in each column per metric is bolded.}
\label{tab:results_id}
\end{table*}

From this initial pool, we filter 40\% of the samples which are potentially noisy and harmful for generalization. In particular, we adapt from the split-and-filter protocol by \citet{gao2025principled}, where we split the data pool into two partitions, train an ensemble of $4$ reward models separately on each partition, and evaluate the the ensemble's per-sample average validation losses on the other held-out partition. Samples in the two partitions are merged and the ones with smallest validation losses are discarded. In our setup, we train lightweight MLP networks as the ensemble models on top of hand-crafted features extracted from each (prompt, response) pair. We used LFTK~\citep{lee-lee-2023-lftk} for extraction of $220$-dimensional features from prompt and resposne each. Discarding low-loss samples according to the ensemble encodes our intuition that those samples potentially carry spurious patterns most easily captured by the shallow features. The  statistics of datasets are shown in Table~\ref{tab:dataset_stats}. We split the filtered set into training and test set by a ratio of $9: 1$, which are shared by LatentRM and all baselines. No hyperparameters are tuned on the test set for all methods, while LatentRM introduces no additional hyperparameter in itself.

\paragraph{Evaluation Benchmarks}
To evaluate OOD generalization, we test on RM-Bench~\citep{liu2025rmbench} and PPE Correctness~\citep{frick2025how}. RM-Bench comprises $4$K preference pairs evaluating an RM's sensitivity to subtle semantic differences and robustness against style bias. PPE Correctness contains $2.5$K challenging prompts requiring deep reasoning, with $5$ preference pairs per prompt. To evaluate listwise candidates, we merge these pairs into deduplicated lists per prompt, score them directly, and convert the predicted rankings back to pairwise decisions~\footnote{Pointwise scalar RM scores each (prompt, response) pair independently for both RM-Bench and PPE Correctness.}. Notably, both benchmarks construct response pairs from the same generating LLM per prompt, closely mirroring real-world RLHF setup and demonstrating high correlation with downstream RLHF performance~\citep{liu2025rmbench, frick2025how}.

\paragraph{Baselines} We evaluate LatentRM against the following baselines:
\begin{itemize}
    \item \textbf{ScalarRM} is a standard pointwise reward model optimized over listwise preferences using Plackett-Luce loss.
    \item \textbf{GenerativeRM} is a listwise generative reward model optimized via RL,  using Kendall's $\tau_b$ as the reward.
    \item \textbf{MultitaskRM} is a generator-discriminator method which uses Kendall's $\tau_b$ reward for the generator and Plackett-Luce loss for scalar RM on reasoning-augmented inputs.     
\end{itemize}

\paragraph{Implementation Details} 
We use \texttt{\textsc{VERL}}~\citep{sheng2025hybridflow} with \texttt{\textsc{vLLM}} rollout engine~\citep{kwon2023efficient} throughout the experiments. We set learning rate to $2\times10^{-6}$ for policy model and $1\times10^{-5}$ for scalar reward model,  maximum prompt length to $16384$. For each prompt, $8$ rollouts are sampled with a temperature of $1.0$ and a maximum response length of $8192$. We set batch size to $128$ and train for $1$ epoch, and adopt pure on-policy learning without PPO-style mini-batching \citep{schulman2017proximal} in case of RL. Token-level KL penalty coefficient is set to $1\times10^{-3}$ for baselines as in \citep{chen2026rmr}. Rewards are transformed to advantages by mean-centering as with LatentRM. Due to cost-intensive online RL training, all experiments are based on a single run with a fixed random seed of $42$.

\begin{table*}[ht]
\centering
\small
\renewcommand{\arraystretch}{1.3}
\setlength{\tabcolsep}{0pt}
\begin{tabular*}{\textwidth}{@{\extracolsep{\fill}} l cccccccc c cccccc @{}}
\toprule
 & \multicolumn{8}{c}{\textbf{RM-bench}} & & \multicolumn{6}{c}{\textbf{PPE Correctness}} \\
\cmidrule(lr){2-9} \cmidrule(lr){11-16}
\textbf{Method} & \textbf{Chat} & \textbf{Math} & \textbf{Code} & \textbf{Safe} & \textbf{Easy} & \textbf{Norm} & \textbf{Hard} & \textbf{Avg.} & & \textbf{MMLU} & \textbf{MATH} & \textbf{GPQA} & \textbf{MBPP+} & \textbf{IF} & \textbf{Avg.} \\
\midrule
ScalarRM           & 71.9 & 71.1 & 65.0 & 94.8 & 86.7 & 78.7 & 61.6 & 75.7 & & 71.0 & 76.1 & 58.4 & 59.5 & \textbf{64.1} & 65.8 \\
GenRM w/o RL       & 69.6 & 85.0 & 65.5 & 90.9 & \textbf{86.9} & 79.7 & 66.7 & 77.8 & & 75.5 & 89.0 & 61.7 & 56.9 & 53.2 & 67.2 \\
GenerativeRM       & 75.0 & \textbf{90.9} & 66.4 & 93.1 & 86.2 & 81.9 & 75.9 & 81.3 & & 73.0 & 85.6 & 59.1 & 50.0 & 52.3 & 64.0 \\
MultitaskRM        & 78.2 & 89.1 & 65.9 & 93.3 & 82.0 & 82.9 & 80.1 & 81.7 & & 80.4 & 92.2 & 69.3 & 61.3 & 56.3 & 71.9 \\
LatentRM           & \textbf{78.5} & \textbf{90.9} & \textbf{66.8} & \textbf{95.1} & 83.5 & \textbf{83.6} & \textbf{81.3} & \textbf{82.8} & & \textbf{80.6} & \textbf{92.7} & \textbf{70.0} & \textbf{61.6} & 55.4 & \textbf{72.1} \\
\bottomrule
\end{tabular*}
\caption{Evaluation results on out-of-distribution RM-Bench and PPE Correctness benchmarks, measured by accuracy. The best value in each column is bolded.}
\label{tab:ood}
\end{table*}

\subsection{Findings on Preference Modeling}

\paragraph{Results on In-distribution Test Sets}

From ID test set evaluation results in Table~\ref{tab:dataset_stats}, we observe that LatentRM outperforms all baselines on general, math, STEM \& Code and adversarial domains, achieving the best log-likelihood and Kendall's tau overall. Importantly, LatentRM wins consistent gains in log-likelihood $\log p_{\varphi}(y|x,z)$ over MultitaskRM except for the safety domain, verifying that the kendall's tau reward used to train state-of-the-art generative RMs \citep{whitehouse2026j} could still be suboptimal when its reasoning serves as an intermediate variable for assisting the scalar RM's decision. Similarly, optimizing the generator for kendall's tau did not lead to the best kendall's tau, compared to LatentRM. LatentRM induces reward signal from the generator to maximize the data likelihood under latent variable model, thus features high fidelity to the scalar RM's goal without any heuristic reward design for the generator.

Compared to the vanilla ScalarRM, both MultitaskRM and LatentRM brings large performance gains which are most pronounced on reasoning-intensive domains such as math, STEM \& Code, and adversarial prompts. This suggests that incorporating deep reasoning in scalar RM could visibly mitigate its reliance on superficial features and boost the performance. On the other hand, generativeRM markedly underperforms scalarRM in Kendall's tau, indicating that scalarRM more effectively fits to the training data distribution. Yet scalar RM tends to overfit the data and suffer from limited generalization on OOD datasets, which we detail below.

\paragraph{Results on Out-of-distribution Datasets} The results in Table~\ref{tab:ood} show that LatentRM maintains the best overall OOD generalization performance, achieving an average of $82.8\%$ on RM-Bench and $72.1\%$  on PPE Correctness.
Importantly, LatentRM wins consistent gains over MultitaskRM on diverse domains and difficulty levels, verifying that our end-to-end learning objective yields reasoning traces that are more likely to transfer to OOD distributions. Meanwhile, scalarRM shows lowest or near-lowest overall performance ($75.7$ on RM-Bench and $65.8$ on PPE), suggesting that its advantage over generative RM in the ID setting no longer holds on OOD datasets. It corroborates evidence from prior works \citep{icml23scaling, miao24inform} that scalarRMs easily suffer from distribution shifts. Nevertheless, LatentRM could still outperform generativeRM, implying that conditioning scalar RM on latent chain-of-thought reasoning notably overcomes overfitting to superficial and in-distribution cues.

From a closer look into each benchmark and its subdomains, LatentRM's improved performance on RM-Bench suggests that LatentRM is capable of discerning subtle variations in content for preference decision, and is robust to stylistic biases that LLM judges are known to suffer from \citep{chen2026rmr}, including tendency to favor long responses or the ones that use markdown format. For instance, LatentRM's gain is most pronounced on \textbf{Hard} subset ($81.3\%$) which poses the hardest challenge on distinguishing stylic biases. We observe such robustness is  uniform across Chat, Math, Code and Safety domains. As for PPE, we see that LatentRM shows markedly better accuracy than GenerativeRM and ScalarRM on reasoning-intensive domains such as MATH ($92.7\%$), GPQA ($70.0\%$), and MBPP+ ($61.6\%$). This suggests that when augmented with reasoning, scalarRM could be much better in preference modeling on reasoning domains than generativeRMs.  

\begin{table}[tbp]
\centering
\small 
\setlength{\tabcolsep}{3.5pt} 
\caption{Comparison of performance with external baselines. Q and L in \textbf{Backbone} column denotes Qwen and Llama, respectively; RM-B: RM-Bench; PPE: PPE Correctness. The overall accuracies are referenced from the original papers. $\dagger$: result reproduced by \citet{liu2025inferencetimescalinggeneralistreward}. }
\label{tab:rm_comparison}
\begin{tabular}{llcc}
\toprule
\textbf{Method} & \textbf{Backbone} & \textbf{RM-B} & \textbf{PPE} \\
\midrule
\multicolumn{4}{l}{\textit{Scalar RMs}} \\
Skywork-v0.2 \citep{liu2024skyworkrewardbagtricksreward} & L3.1-8B-it & 64.7 & 62.5 \\
SteerLM-RM \citep{wang24helpsteer2}  & L3.1-70B-it  & 72.2 & 63.2 \\
\midrule
\multicolumn{4}{l}{\textit{Generative RMs}} \\
RM-R1 \citep{chen2026rmr}  & Q2.5-7B-it  & 70.2 & -    \\
RM-R1  \citep{chen2026rmr}  & Q2.5-14B-it & 76.1 & -    \\
J1 \citep{whitehouse2026j}     & L3.1-8B-it  & 73.4 & 52.8 \\
J1  \citep{whitehouse2026j}    & L3.3-70B-it  & 82.7 & 70.2 \\
PaTaRM \citep{26patarm} & Q3-8B           & 78.7 & -    \\
\midrule
\multicolumn{4}{l}{\textit{reasoning-driven scalar RMs}} \\
CLoud$^\dagger$ \citep{ankner2024critique}   & Gemma2-27B              & -    & 62.4 \\
LatentRM & Q3-4B-it       & 82.8 & 72.1 \\
\bottomrule
\end{tabular}
\end{table}

\paragraph{Comparison with External Baseline Models}
We report comparisons against external reward models in Table~\ref{tab:rm_comparison}. Since they employ different data curation strategies and pre-trained model backbones, the reported numbers are not directly comparable to our baselines and serve primarily as a reference point. In general, we find LatentRM achieves favorable performance that matches or outperforms much larger generative and scalar models (e.g., 27B–70B parameters) across both RM-Bench and PPE Correctness benchmarks. Closest to our approach, for example, CLoud \citep{ankner2024critique} achieves an overall accuracy of 62.4 on PPE. 

\begin{figure*}[t]
    \centering
    \includegraphics[width=\linewidth]{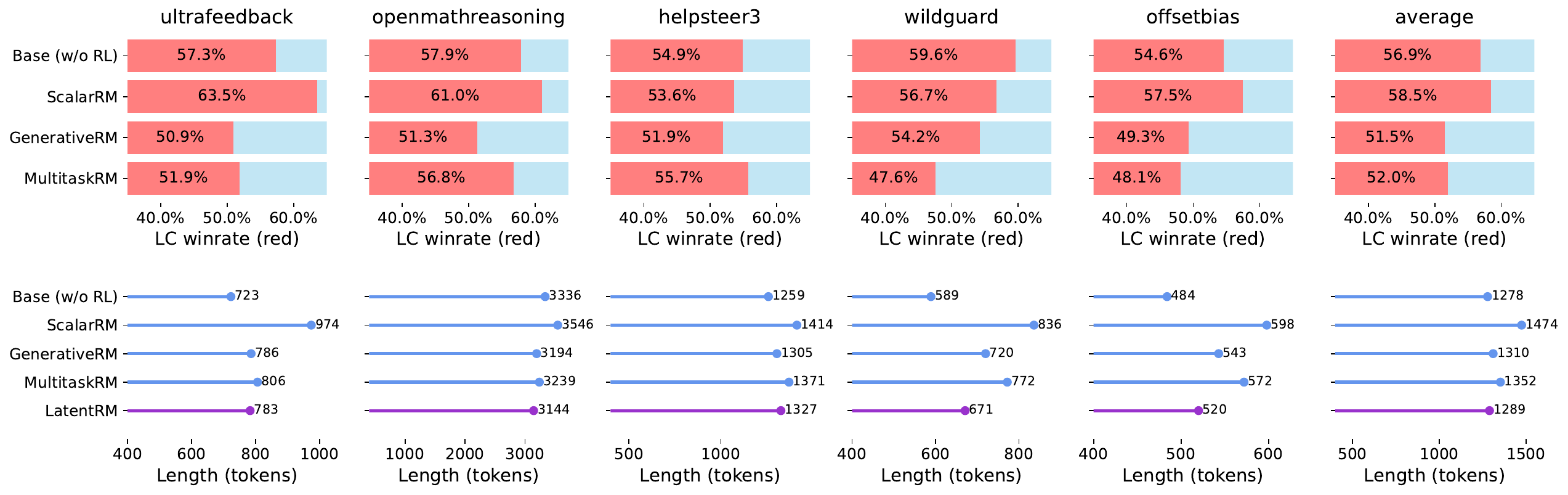}
    \caption{Length-controlled (LC) win rate and token length of post-RLHF policy rewarded by \textbf{LatentRM} against post-RLHF policy rewarded by baseline RMs, evaluated using Qwen3.7-plus \citep{qwen37plus} as judge. We used \citet{dubois2024lengthcontrolled} for the judge template and LC winrate computation formula which corrects the Judge's length bias from the winrate by using a regression model. }
    \label{fig:rlhf}
\end{figure*}

\subsection{Findings on RLHF}

To evaluate LatentRM and baselines in online alignment, we perform $100$ steps of online RLHF using GRPO \citep{shao2024deepseekmath} on Qwen3-4B-Instruct as policy model. We set KL-loss coefficient of $10^{-2}$, a batch size of $128$, $4$ rollouts per sample, and maximum token length of $4096$ for both prompt and response. After training policy models under different RMs, we randomly sample $200$ samples from each subset of the test dataset, computing winrates of the policy by LatentRM against the policy by other RMs. The results are shown in Figure~\ref{fig:rlhf}.

Figure~\ref{fig:rlhf} shows that the RLHF policy guided by LatentRM yields average length-controlled (LC) winrates exceeeding 50\% against the base policy ($56.9\%$), and RLHF policies by scalar RM ($58.5\%$), Generative RM ($51.5\%$) and MultitaskRM ($52.0\%$). These gains demonstrate that LatentRM's offline preference modeling strengths (Table~\ref{tab:results_id}, \ref{tab:ood}) effectively generalize to dynamic, non-stationary policy shifts. Notably, LatentRM achieves these results without verbosity exploitation, generating the shortest average outputs (1,289 tokens) among all RLHF baselines. Conversely, the Scalar RM policy suffers from reward overoptimization, yielding lower win rates against LatentRM than even the base policy, highlighting the necessity of explicit reasoning traces for stable online reward guidance.

Across the tested domains, LatentRM shows consistent benefits over the baselines on general open-ended conversation (Ultrafeedback), mathematics (OpenMathReasoning) and STEM \& Code (HelpSteer3) domains. The clear advances on reasoning domains signify that LatentRM develops genuinely informative reasoning traces. Yet on safety (WildGuard) or adversarial (OffsetBias) domain, LatentRM lags behind MultitaskRM, which partially aligns with the ID results in Table~\ref{tab:results_id}. We hypothesize that as UltraFeedback and OpenMathReasoning make up $70\%$ of the training dataset, LatentRM devotes more optimization efforts on these representative domains at certain cost of performance on underrepresented domains.  

\begin{figure}[t]
    \centering
    \includegraphics[width=\linewidth]{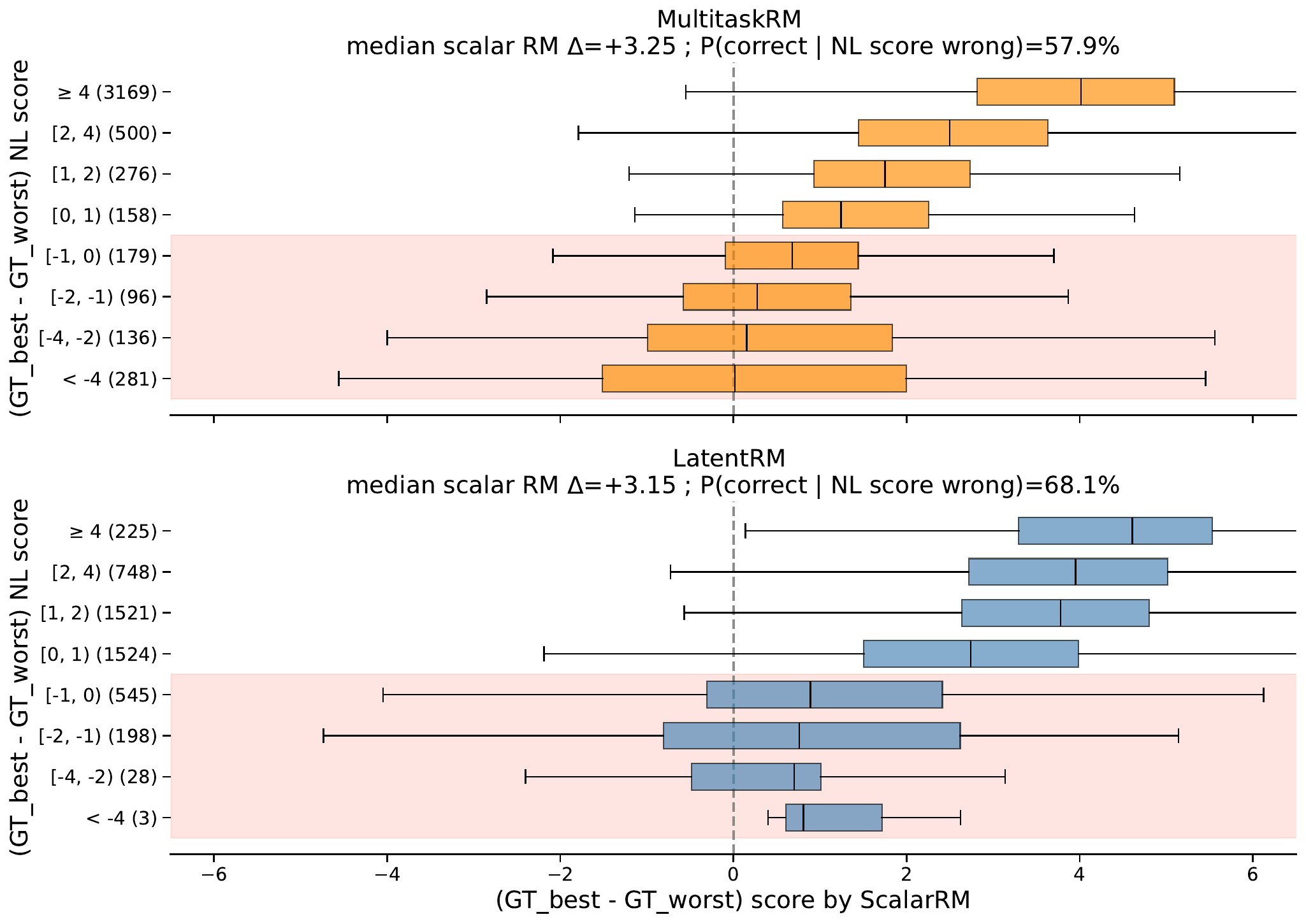}
    \caption{Per-sample score gaps $\Delta$ (best response $-$ worst response): generator's natural language score $\Delta$ bins (y-axis) vs. downstream Scalar RM $\Delta$ (x-axis). Red band denotes regions where natural language scores disagree with $y$. The distributions for the boxes are aggregated over our ID test sets. }
    \label{fig:detal_dist}
\end{figure}

\subsection{Analysis of Score Gap Distribution}
To draw deeper insights into the internal mechanism of LatentRM compared to MultitaskRM, in Figure~\ref{fig:detal_dist} we visualize the distribution of per-sample score gaps between the best response and the worst response according to $y$, jointly over the generator's natural language scores (y-axis) and the downstream scalar RM's scores (x-axis). When the natural langauge scores align with the groundtruth ranking, the corresponding scalarRM produces large positive score gaps, yet LatentRM's scores provides clearer distinction than MultitaskRM. Moreover, when the natural language scores are incorrect (the red regions in the figure), LatentRM exhibits exceptional recovery to correct decision by the downstream scalarRM. In contrast, MultitaskRM tends to recover much less from such cases. This suggests that LatentRM does not simply echo the verbalized scores for its decision, but actively leverages the intermediate reasoning traces as supporting evidence. We hypothesize that this is a positive effect of end-to-end training, where the generator focuses more on developing informative reasoning trace without pressure to yield correct natural language scores.

\section{Conclusion}
In this paper, we introduce LatentRM, a scalar reward model enhanced by a reasoner that learns to shape its reasoning trace for a unified goal of maximizing the scalarRM's likelihood of observing groundtruth preference ranking. By casting reasoning trace as a latent variable in a conditional generative model, we derive a simple and natural end-to-end procedure for training LatentRM. Extensive validations on ID and OOD datasets as well as RLHF experiments  demonstrate that LatentRM achieves overall consistent improvements in preference modeling and policy alignment over a strong multi-task learning baseline, and much more clearly over scalar and generative reward models in isolation.

\appendix

\bibliography{aaai2027}

\section{Full Prompt Template of Listwise Generator}
\label{apdx:full_template}

\begin{figure*}[t]
    \centering
    \includegraphics[width=\linewidth]{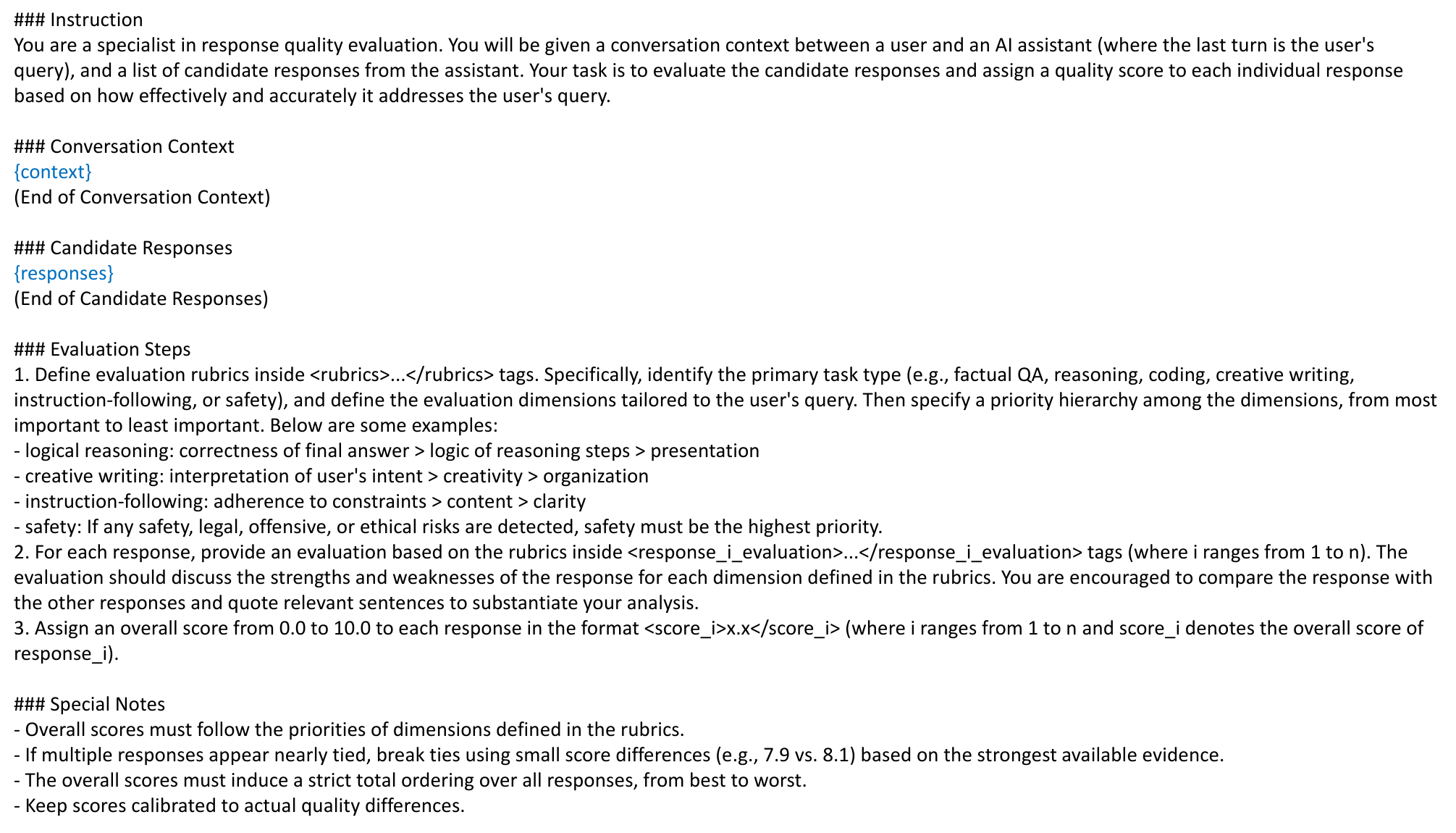}
    \caption{The complete prompt template of our listwise generator. Brackets in blue denotes contents to be filled with conversation context or candidate responses.}
    \label{fig:full_template}
\end{figure*}

We present the full template of our listwise generator in Figure~\ref{fig:full_template}. This prompt template instructs an LLM generator to evaluate the candidate responses by guiding the generator through a structured three-step process: (1) defining task-specific rubrics with a priority hierarchy, (2) writing detailed qualitative analysis per response using pre-defined tags, and (3) assigning a final numerical score from 0.0 to 10.0 that enforces strict total ordering across all candidates.

\end{document}